# What Neuroscience Can Teach AI About Learning in Continuously Changing Environments

Daniel Durstewitz[1,2,3,*], Bruno Averbeck[4], Georgia Koppe[2,5]

[1]Dept. of Theoretical Neuroscience, Central Institute of Mental Health, Medical Faculty, Heidelberg University, Germany
[2]Interdisciplinary Center for Scientific Computing, Heidelberg University, Germany
[3]Faculty of Physics and Astronomy, Heidelberg University, Germany
[4]Section on Learning and Decision Making, National Institute of Mental Health, Bethesda, USA
[5]Hector Institute for AI in Psychiatry, Central Institute of Mental Health, Medical Faculty, Heidelberg University, Germany
[*]Corresponding author (daniel.durstewitz@zi-mannheim.de)

## Abstract

Modern AI models, such as large language models, are usually trained once on a huge corpus of data, potentially fine-tuned for a specific task, and then deployed with fixed parameters. Their training is costly, slow, and gradual, requiring billions of repetitions. In stark contrast, animals continuously adapt to the ever-changing contingencies in their environments. This is particularly important for social species, where behavioral policies and reward outcomes may frequently change in interaction with peers. The underlying computational processes are often marked by rapid shifts in an animal's behaviour and rather sudden transitions in neuronal population activity. Such computational capacities are of growing importance for AI systems operating in the real world, like those guiding robots or autonomous vehicles, or for agentic AI interacting with humans online. Can AI learn from neuroscience? This Perspective explores this question, integrating the literature on continual and in-context learning in AI with the neuroscience of learning on behavioral tasks with shifting rules, reward probabilities, or outcomes. We will outline an agenda for how specifically insights from neuroscience may inform current developments in AI in this area, and – vice versa – what neuroscience may learn from AI, contributing to the evolving field of NeuroAI.

## Introduction

Natural environments are never constant. Food sources may be depleted; novel, even more nutritious opportunities may pop up. Strategies for escaping from harmful situations that worked in one context may fail in another. Changing weather conditions, seasons, or migration into novel terrain may impose new challenges. This is even more true in interaction with predators, potential prey, and conspecifics, who may change their own behavioral policies in response to yours, leading to multi-level feedback loops. Animal brains evolved over hundreds of millions of years to deal with such challenges. Oftentimes, behavioral strategies have to be efficient and fast, otherwise they may leave their enactor behind dead. You only get one chance to escape from a shark tracking you, you cannot repeat the experiment and hope for a better outcome. This ability to continuously adapt to changing environmental conditions and to come up with new solutions on the fly, especially in dangerous and time-critical situations, is a characteristic of increasing importance to AI. It is of obvious relevance to robots and autonomous agents directly interacting, like animals, with the physical world. It, however, also plays an increasing role for AI systems engaging with humans online, like medical support systems or other large language model (LLM) based assistants, leading toward 'agentic AI'.[1]

The process of training current state-of-the-art (SOTA) machine learning (ML) and AI models fundamentally differs, however, from how animals learn. Unlike animals, they are usually trained once on a huge corpus of data. After this initial training, there is comparatively little flexibility in readjusting parameters to cope with novel scenarios. The training itself is slow and gradual, commonly based on forms of gradient descent (GD). This is strikingly different from animal brains which continuously absorb new knowledge, and may ingrain one-time-only experiences forever within the physical structure of their synaptic connectivity.[2,3] Although recent

LLMs like DeepSeek R1 or OpenAI's o1 showcase impressive reasoning skills, the versatility and flexibility with which small animal brains adapt to non-stationary environments is still unmet by today's most advanced AI systems.

Here we will explore how mechanisms discussed in (computational) neuroscience on how animal brains deal with non-stationary environments may inspire AI research. We will start with a brief overview of the ways in which current AI models adjust to novel tasks and statistical shifts. This includes strategies for continual and lifelong learning in ML, and the more recent phenomenon of in-context learning. We will then summarize important behavioral and neurophysiological observations on non-stationary tasks in animal research, which are similar in design to those used for probing continual and in-context learning in AI. Two topics in (computational) neuroscience are particularly relevant for explaining behavioral performance on such tasks. These are neuro-dynamical mechanisms and synaptic mechanisms. We will draw attention to how these mechanisms in animal brains crucially differ from current AI algorithms, and how they reflect adaptation to the multiple demands and timescales of their non-stationary environments, thus illuminating their potential for AI. We will conclude with a discussion of how knowledge from neuroscience could be transferred into AI development, and how – vice versa – progress in ML/AI may fertilize and guide neuroscience.

## Learning in non-stationary environments in AI systems

Fundamentally, there are two ways of dealing with novel situations that profoundly differ from what an agent has experienced before (referred to as out-of-domain or out-of-distribution (OOD)). It can adapt its 'parameters' to incorporate these new experiences, sometimes dubbed "in-weights" learning.[4,5] In ML/AI, this includes any form of parameter fine-tuning and algorithms for continual and lifelong learning. Or it could try to infer online from the observations the optimal response, related to what has been called "in-context" learning,[6-9] which includes methods of prompting to induce reasoning processes, like chain-of-thought.[10] We will start by reviewing these two areas in ML/AI.

### *Adapting to changing environments through continual learning*

AI systems are usually trained through iterative numerical schemes, most commonly some form of GD. The training algorithm is given some loss function $\mathcal{L}(\boldsymbol{\theta}; \boldsymbol{X})$, such as a mean-squared error loss or a (negative) likelihood, where $\boldsymbol{X}$ are the data and $\boldsymbol{\theta}$ the model parameters. GD then leads an initial parameter guess $\boldsymbol{\theta}_0$ downhill, recursively improving it by moving against the gradient with $-\gamma \nabla \mathcal{L}(\boldsymbol{\theta}; \boldsymbol{X})$, where $\gamma$ is a learning rate. Common optimizers like Stochastic GD (SGD) or RAdam[11] refine this process by injecting stochasticity or using adaptive learning rates.

GD, like essentially any numerical optimization scheme, is a slow gradual procedure that requires – depending on model and data size – up to billions of repetitions and training samples for the process to converge.[7,12] It is computationally costly and data intense. Retraining a full model as more data become available is computationally prohibitive. Moreover, there is the problem of *catastrophic forgetting*:[13-15] As a model is retrained on new tasks, it tends to forget previous solutions, which is fatal if old tasks reappear in the model's environment (Fig. 1a). The problem of maintaining adaptivity while preserving previous skills has also been dubbed the 'plasticity-stability-dilemma'.[16,17] It has been a known issue for decades, almost since the serious study of neural networks took off in the 80s.[16,18,19] Numerous ways of addressing it have been suggested, essentially falling into four major categories[20,21] (excellent previous reviews of this field, including connections to biology, are available[22-24], so we will be brief here): First are solutions that prohibit strong shifts in network parameters in order to prevent catastrophic forgetting;[13,25,26] second, architectural solutions like modularity or freezing part of the structure in later learning;[27-30] third, replay of previous content interspersed with new tasks;[31,32] fourth, functional solutions like partial network resets to maintain plasticity.[33] To test the capability for improved continual learning, certain benchmarks were defined in the field. Fundamentally, these consist of an initial training on a set of tasks, and then continual training on new tasks with task identity provided (*task-incremental* *learning*), new domains or contexts (*domain-incremental learning*), or new classes without identifying the switches between different tasks as such (*class-incremental learning*).[21,34]

Regarding the first approach, the idea is to penalize strong shifts in parameter space which would erase previously learned tasks. This could be implemented through regularization terms that keep updated parameter estimates close to previously successful solutions or within a Bayesian setting which places a strong prior on previous parameters.[13,25,26] Regarding the second approach, a standard strategy for fine-tuning LLMs on new data is keeping part of the system frozen and only allowing changes to some of the parameters. This can happen in various ways, e.g. only making the final layers of a transformer network trainable,[35] or through "low-rank adaptation (LoRA)" which decomposes weight matrices into fixed (previously trained) and trainable parts.[36] Another strategy that particularly helps transfer to new domains is modularity, with different network modules learning different subtasks.[28,37] Experience replay, the third class of approaches, is a procedure inspired by the phenomenon of replay first observed in the rodent hippocampus, where during sleep (but also awake states, as we know now) previously visited sequences are re-instantiated in neural activity.[38,39] Experience replay schemes mix new task learning with selective presentation of old class samples in an optimal way.[31,32] An example of the fourth strategy is the recent work by Dohare et al.[33] on continual back-propagation (Fig. 1a): Here, a small percentage of network units with only little contribution to the activity of other units is continuously reset, i.e. re-initialized with new weights, keeping the network alive. This is based on the insight that during continual learning 'unit debris' starts to pile up, i.e. many units saturate and no longer contribute, thus reducing the flexibility of the system to adapt to new tasks.

While all these methods reduce or – more or less – abolish the catastrophic forgetting problem, they are mostly still based on GD training and hence inherit its natural limitations: Learning is slow, gradual, and costly, requiring many repetitions and a large number of samples. Moreover, the focus of most of these methods is on eradicating catastrophic loss of previous memories. Only few approaches, like refresh learning,[20] have shifted attention toward embracing new information. Either way, updating is usually neither rapid nor satisfied by just a few samples, in stark contrast to biological learning.[3,40] Rather, in modern AI, anything based on few samples (few- or zero-shot learning) or quick updating is left to online inference, discussed next.

*Adapting to changing environments through inference*

LLMs and LLM-based reasoning models show an apparently remarkable capability for OOD generalization. The term OOD most commonly refers to shifts in the statistical distribution of the test set[41] (in contrast to the classical statistical out-of-sample generalization[42]), but may also refer to other strong shifts in tasks or conditions, like changes in the dynamical regime a time series comes from.[43] A particular form of few-shot OOD generalization in LLMs has been dubbed "in-context learning (ICL)".[7,9] The term "learning" might be slightly confusing here, as there is actually *no learning in terms of parameter re-training or fine-tuning*. Rather, the model is presented with a sequence of samples $\{x_i, y_i\}$, $i = 1 \dots N$, from a new task or function class not contained in the model's training set. It then infers the correct answer on a probe item $x_{N+1}$ (Fig. 1b). Zero-shot refers to the case "$N = 0$", i.e. if the model is able to infer the correct output directly for a new task just from the instruction or type of query, as in reasoning processes elicited by 'chain-of-thought (CoT)' prompts.[10,44]

Since in massively pre-trained LLMs this phenomenon is hard to analyze, partly because properties of the training corpus are not always clear, recent studies tried to isolate it by training transformers or other foundation architectures from scratch.[4,9,45,46] For instance, in Garg et al.,[9] transformers were trained on a set of regression problems $\{x_{i,k}, y_{i,k} = f_k(x_{i,k})\}, i = 1 \dots N, k = 1 \dots K$, and then during test time were shown a sequence of samples $\{x_{i,l}, y_{i,l} = f_l(x_{i,l})\}, i = 1 \dots M$, from a new function $f_l$ not contained in the training set, i.e. $\forall k: k \neq l$ (Fig. 1b). The mechanisms underlying ICL are still not well understood. A strong claim here is that ICL rests on a kind of 'in-context GD', based on proofs by construction that successive transformer layers could implement GD steps.[6,47] However, this idea relies on strong assumptions about the model's weight matrices which are empirically not met, nor do the output distributions produced by GD and ICL agree.[48-50] Also, it is hard to conceive how millions of GD steps needed for more challenging tasks could be implemented this way. A perhaps more mundane, but at the same time more biology-friendly explanation could be that ICL simply relies on retrieving close examples through a process resembling recall in associative memory.[51] Other hypotheses are that ICL interpolates among training samples,[52] utilizes compositional structure,[53] or recombines tasks learned in pre-training.[49]

With in-context GD an unlikely explanation for the observed ICL capabilities, ICL remains limited by properties of the training distribution.[52,54] Training sets therefore need to be massive, and new data or tasks, violating the training distribution, seem hard to accommodate. Moreover, even test-time compute can be expensive and comparatively slow.[55] The brain has found other solutions, as we will illuminate in the following, where new learning and inference are often tightly interwoven.

**Learning in non-stationary environments in animals**

Many tasks in neuroscience are close in design to those used for benchmarking AI systems, in the sense that consecutive samples of "input/ output pairs $\{x_i, y_i\}$, $i = 1 \ldots N$" (trials) are presented. The inputs are commonly (but not necessarily) discrete sets of stimuli, and the outputs are requested behavioral choices. However, the correct output is rarely explicitly conveyed to the animal, but only indicated through a reward signal, making most tasks in neuroscience reinforcement learning paradigms. An (operant) rule defines the correct mapping from inputs to outputs. In some tasks this mapping is probabilistic, mimicking the uncertainty in the world. There are several types of *non-stationary* task paradigms in use that involve shifting options, rules, or reward values, some of which – and their basic homology to AI benchmarks – we will briefly review first.

In rule learning or shifting tasks (Fig. 2a), animals are first trained to perform on one rule, and after achieving a certain performance level, task rules (reward contingencies) are changed unbeknownst to the animal. Rule shifting tasks are an example of what would be considered a continual learning paradigm in ML/AI, with novel rules introduced consecutively, but old rules repeated once in a while.[56,57] They are more in line with class- rather than task-incremental learning, since task identity is not provided to the animal – it must be inferred by sampling from the environment. Consequently, as important in natural environments, *exploration* plays a huge role in the animal's behavior. There are also paradigms with explicit context cues indicating the active rule, more similar to task-incremental learning.[58] Context is important for behavioral flexibility, allowing an animal to adjust its behavior depending on other environmental variables.[58-60]

If one focuses on a particular rule switch, one may also see rule shifting as an ICL paradigm: Given a sequence of observed trial outcomes, the animal needs to infer the new task rule. In ML/AI, continual learning depends on parameter changes, while ICL is an inference process not requiring in-weights learning. In animals, there is no clear distinction, and behavior on rule shifting tasks may involve either one or a mixture of both. The exact tradeoff may depend on an animal's previous exposure to various rules and its natural biases.[57] Thus, the neural mechanisms may differ depending on whether rules are learned *de-novo*, or whether they are already in the animal's repertoire and just revisited.[57,61] This, however, may not be immediately evident from the environmental input.

Other task paradigms that involve continuous changes in the animal's environment are drifting n-armed bandit and gambling tasks.[62-64] In these tasks animals or humans are confronted with a set of choice options. After choosing an option a reward is delivered. Different choice options vary in the size or probability of reward, and the task is to learn which option leads to the largest reward. Learning is engaged by periodically changing the choice options in various ways. For instance, the reward value associated with each option can vary over time, as exemplified by experiments with rodents in Passecker et al.[63]: Here, rats could make a choice between two options in a Y-arm maze, one yielding a low probability high reward, and the other a high probability low reward. After a block of trials, reward magnitude or probability of the options was altered, without this change being explicitly signaled to the animals. In another experiment by Tang et al.[62] with monkeys, new choice options were introduced periodically, such that the animals needed to choose between exploring the new, potentially more rewarding option, or exploiting the known options. Hence, all of these task manipulations require subjects to update their choice policies continuously. One difference to rule learning and shifting tasks is that the rules of the game do not change, and this restricts the solution space.

A couple of observations about the neuronal and behavioral processes in such tasks are potentially relevant to AI: First, even if a rule is completely new, it rarely takes an animal thousands of repetitions to learn it, but often a couple of trials are sufficient. Second, animals continue to explore other options even when the task rules are stable.[65] This is a reasonable prior in *non-stationary* environments where reward contingencies may change once in a while. Third, performance relies on previously learned task elements, rules, natural biases, task schemata, and concepts.[29,66-68] Procedures for animal training in the lab usually consist of several stages of familiarizing the animal with the different task elements ("shaping"),[65] reminiscent of curriculum learning protocols in ML/AI. Fourth, behavior does not change gradually throughout the task, but often there are sudden performance jumps (Fig. 2b).[69,70] These jumps in behavioral performance are accompanied by rapid transitions in neuronal representations in prefrontal cortex and other brain areas (Fig. 2c).[57,71,72] Such jumps are observed even if a rule is novel and has not previously (explicitly) been introduced to an animal.[71] Rapid performance and neuronal changes also mark the onset of exploratory phases, when an animal realizes that a previous behavioral strategy no longer works.[71,73] Fifth, in tasks where a previous response is extinguished ("extinction learning"), i.e. is no longer rewarded, the previous behavior is not unlearned nor forgotten, but suppressed.[65] For instance, extinguished behaviors can be rapidly reinstated once the reward is reintroduced.[65] Interestingly, extinction learning, like acquisition of a new rule, goes hand in hand with sudden transitions in behavior and neuronal population activity.[74]

These are all properties which are not easily reconciled with the training and behavior of neural networks in ML, but which are fundamental to our understanding of how animals quickly adapt in the wild. Two classes of mechanisms may support these functions in animal brains, *dynamical* and *plasticity* mechanisms, which we will discuss in turn.

**Neuro-dynamical mechanisms supporting adaptation to non-stationary environments**
Dynamical systems theory (DST) provides a general mathematical framework for understanding the behavior of natural or engineered systems evolving in time, formalized as systems of differential equations or recursive maps. DST has been a mainstay in computational and theoretical neuroscience for decades, serving well for explaining computation in the nervous system and how it evolves from the underlying physiology. Dynamical systems are Turing complete.[75-77] While DST is increasingly recognized also in ML/AI as a toolbox for explaining the behavior of transformers,[78,79] RNNs,[80,81] or training algorithms like SGD,[82-85] how dynamics could be utilized to perform computation has been less appreciated in the AI field. Notable exceptions are Hopfield networks[86] and Boltzmann machines,[87,88] whose attractor dynamics support robust memory storage, retrieval, and pattern completion. Diffusion models exhibit similar dynamical mechanisms.[89,90] The perspective DST provides on computation in the brain could be of major relevance for agentic AI and online learning in natural environments, partly because it puts the emphasis on the temporal aspects and many timescales on which biological and physical processes unravel.

DST deals with the topological and geometrical properties of state spaces (Fig. 3), subsets of Euclidean space which contain all states a dynamical system could be in (i.e., the set of values the dynamical variables can attain). The temporal evolution of a system corresponds to a trajectory through state space (Fig. 3), potentially – in the infinite time limit – converging to geometrical objects called attractors, which could be single points (fixed points/ equilibria), limit cycles (corresponding to stable oscillations), or objects with more complicated, fractal geometry, called chaotic attractors. Computational neuroscience conceives computational processes as being implemented in terms of such trajectories, with their fate ruled by the state space's geometrical properties. There is substantial experimental support for the existence of various types of attractor objects in the brain and their role in computation.[91-93]

One important concept is that of a manifold attractor,[94-98] a continuous set of marginally stable equilibria toward which the system's state evolves from various initial conditions, while on the manifold itself there is no further movement unless the system is pushed by external inputs or noise (Fig. 3a). A manifold attractor provides a computational template for "long short-term" (working) memory: External inputs guide the system's trajectory to a unique state on the manifold, storing that information indefinitely if unperturbed (Fig. 3a). This is in the absence of any parameter changes. Empirical evidence for manifold attractors in the brain abounds.[91-93,99] One classical example is the ability of fish and mammals to maintain arbitrary eye positions.[91,95,100] Physiologically this seems to be supported by a nearly continuous set of stable neural firing rates, evident even at the single cell level.[101] Another example is the nearly continuous representation of place or head direction in the rodent hippocampus.[92,94] In RNNs, manifold attractors can be encouraged through regularization terms in the loss function,[102] leading to systems that outperform other architectures, like LSTMs,[103] on typical long-range arena problems. Manifold attractors can also integrate continuous variables like reward feedback in a bandit task, thus adapting the system to changing environments without requiring synaptic plasticity.[64] In some sense, manifold attractors are a dynamical substrate for a long context window, yet potentially much less resource-intensive than context windows in transformer-based models.

The brain may engage many much more complex dynamical constructs to realize computational properties, and even computational neuroscience has hardly scratched the surface here. Separatrix cycles,[104] or heteroclinic channels,[105,106] are sets of saddle nodes (or cycles) connected through so-called heteroclinic orbits that lead from one node to the next. These allow for robust yet flexible implementation of sequences, in which elements can quickly be exchanged. Hence, they also enable fast adaptation to novel contexts, which, for instance, require novel structuring of a series of motor commands.[106] Relatedly, chains of ghost attractors have been proposed to serve a similar purpose.[107-109] These are remnants of attractors which have lost their stability, so are not formally attractors anymore, but still pull in trajectories from a larger surrounding which then slow down as they move into the ghost's immediate vicinity (Fig. 3a, right; Fig. 3c). Neural activity observed in the rat anterior cingulate cortex during multiple-item working memory and decision making tasks has properties consistent with such ghost attractor chains, with trajectories considerably slowing down as animals reach decision points in the task (Fig. 3d).[110] Attractor ghosts can also explain the observation of ramping neuronal activity profiles with adjustable slopes in timing tasks (Fig. 3a,b),[96,111,172] and have been implemented into RNNs for them to capture arbitrary time scales.[102] The idea of chaotic itinerancy, the chaotic wandering among ghost attractors, takes this one step further and has been suggested to underlie flexible and adaptive cognitive operations.[108] A crucial take-home here is that these dynamical mechanisms not only allow for flexible and fast reconfiguration in novel surroundings, but also involve a natural notion of time and timescales absent in modern AI systems: The world's non-stationary evolution is mirrored by the brain's non-stationary dynamics.[112]

Another important concept in DST is that of a bifurcation, related to phase transitions in physics or tipping points in many complex natural, e.g. climate, systems. A bifurcation is a qualitative, topological change in a system's state space as one or more of its parameters are changed.[104] Many types of bifurcations incur abrupt changes in the system dynamics as one of the attractor objects suddenly vanishes and/or novel objects appear. Bifurcations may drive the restructuring of dynamical motifs supporting different behavioral rules,[67] accounting for the sudden jumps observed in neuronal representations and behavior in the rule shifting tasks discussed above.[71] Interestingly, such sudden jumps also occur during GD-based training of recursive or auto-regressive AI models, where they empirically and formally could be linked to bifurcations.[85,113] Systems harboring ghost attractors are close to multiple bifurcations in parameter space, and, more generally, there is some evidence that brains operate in near-bifurcation regimes.[114-117] This may explain their versatility in rapidly adjusting to novel situations: Even fairly small changes in parameters could lead to a fundamental reorganization of the system's computational structure for a novel purpose at hand.

In conclusion, the dynamical systems perspective so important in theoretical neuroscience, could also help improve AI architectures and training algorithms. At a methodological level, it offers a powerful mathematical language for analyzing and understanding the inner workings of AI models and comparing them to brain mechanisms,[118] allowing for the translation between both areas within a common formal framework. At a computational level, insights about how the brain implements flexible and versatile real-world computations in dynamical terms could be transferred into AI models, for instance by suitable regularization priors.[102]

## Plasticity mechanisms supporting adaptation to non-stationary environments

In the brain, changes in synaptic function, but also cellular changes in the contribution of different voltage-gated ion channels to neuronal activity,[119] are thought to underlie learning and behavioral adaptation.[2] There are multiple mechanisms and forms of synaptic plasticity, operating on many different time scales, with differential expression across brain areas.[2,120] There is synaptic short-term depression, facilitation, post-tetanic potentiation, and augmentation, within hundreds of milliseconds to possibly minutes. Forms of long-term depression (LTD) and potentiation (LTP) may set in within minutes, and last up to hours, days, or longer.[2,120] Synaptic plasticity may involve structural changes, like changes in the morphology of synaptic boutons, dendritic spines and possibly arbors, the formation of new or the removal of existing synapses.[121] Surprisingly, even such structural changes may set in within minutes to hours during or after a learning experience.[122,123] Many forms of longer-term plasticity depend on the precise spiking (action potential) patterns among pre- and post-synaptic neurons, so-called spike-timing-dependent plasticity (STDP).[2] Global synaptic scaling rules, similar to weight normalization in ML/AI, ensure that neuronal activity does not run wild, as in epilepsy.[120,124]

Relating synaptic plasticity to in-weights learning, i.e. parameter changes in ML/AI models, there are a couple of noteworthy observations: First, the many forms and types of synaptic plasticity once again highlight that in biological systems timing and time scales are crucial – physical, biological, and societal systems dynamically evolve in time, across many different scales, a feature reflected in physiological priors. Second, since synaptic plasticity often depends on the precise order of pre- and postsynaptic spikes, there is a sense of inbuilt causality to it: Asymmetry in spike-time-dependency may help to ingrain predictive relationships between environmental events.[125] Causality is obviously important for biological beings interacting with the physical world. The fact that even at this fundamental level of connectivity changes principles of causality may play a role, unlike in AI model training, is probably important for design of agentic AI. Third, synaptic plasticity is *ongoing and continual*. There are no separate training and testing or deployment phases. Fourth, a lot of plasticity in the nervous system is *unsupervised* and *local*, like STDP, although semi-supervised or supervised forms of plasticity, involving forms of back-propagation, exist as well.[126,127] This may allow for considerable speed-ups and less resources in training, as no global error signals need to be propagated through the whole system. Fifth, synaptic changes can be rapid[122,123] – in fact, they could happen within just one or a few trials,[2,128] i.e. implement a form of one-shot learning, a point discussed in more detail below. Sixth, plasticity is itself subject to modulation by other factors, so-called meta-plasticity, a phenomenon which has been discussed for its potential role in preventing catastrophic forgetting[129,130] and supporting the formation of schemata.[131]

Another important observation for continual and few-shot learning is that different brain areas are differentially engaged in various forms and stages of learning and memory. This had led to the hypothesis of *complementary learning systems* as a means to prevent catastrophic forgetting.[60,132,133] In particular the hippocampus plays a crucial role here. It is the brain structure best characterized from the cellular and synaptic up to the behavioral levels in terms of plasticity, learning, and memory.[2] One basic idea is that the hippocampus functions as fast memory buffer that over the course of a day (or night, depending on which type of animal you are) collects experiences and integrates them into *episodic memory*.[134,135] This remarkable ability of animals and humans to store away one-time-only experiences on the fly and recall details of it days, weeks, or even years later, even in the absence of explicit positive or negative reinforcers, has been termed *latent or incidental learning*.[136,137] Latent learning and episodic memory, potentially in conjunction with experience replay, may still support reinforcement learning (RL), however, by allowing RL to solve the credit assignment problem across potentially long temporal gaps:[138] Details that seem innocuous now may become relevant later.

Daytime experiences are then replayed to the neocortex during sleep[139,140] for careful integration into existing memory structures and schemata.[68,133] Replay in hippocampus refers to the almost exact repetition of spiking patterns across multiple neurons that accompanied, for instance, sequential traversal of different places as part of a previous experience.[39,140] This replay of previous patterns happens, however, at a compressed timescale in sleep.[140] Complementary learning and memory systems might be set up to minimize the risk of overwriting precious memory contents while serving fast updating, and for balancing a need for detail with the formation of abstract, generalizable concepts.[68,133,141] While experience replay is a technique already utilized in continual learning,[31,32] the idea of different learning systems has received comparatively little attention in modern AI. Given that incidental learning and episodic memory is so fundamental for continual learning in real-world environments for animals, the brain's organizational principles and plasticity mechanisms here are likely to be a rich source of inspiration for agentic AI.

A key aspect here is that this form of learning is *rapid and unsupervised*, with a possible cellular substrate for it identified recently – *behavioral time scale plasticity* (BTSP).[2,40,128] BTSP, first examined in hippocampal place cells, sets in within seconds and requires only one or a few expositions to create a lasting memory of an environmental event. Computational studies found that the physiological principles of BTSP can indeed support rapid formation of associative memories with cue-dependent recall, which are also more robust to overwriting than other mechanisms (Fig. 4).[3,142] It is conceivable that such rapid forms of synaptic plasticity also provide mechanisms for in-context learning, updating task representations after only a few presented examples, a potential largely unexplored in modern AI. The plethora of plasticity time scales[143] in the nervous system may enable forms of adaptivity to changing environments not even touched upon yet by AI. For instance, even classical short-term forms of plasticity, like short-term facilitation and depression, may quickly alter the dynamical landscape of cortical networks operating in near-bifurcation, critical regimes (see previous sect.). The arrangement of ghost attractors could thus be adapted on the fly as task demands change, possibly explaining performance jumps in rule learning tasks.

Relatedly, the potential for plasticity itself undergoes massive changes throughout a lifetime, following specific developmental trajectories. While all cortical areas retain a capacity for plasticity throughout life, in primary sensory and motor systems the developmental window closes early, with cognitive systems following later.[144,145] Thus, basic sensory-motor processes are learned early and stabilized before more complex skills are acquired. This allows the brain to efficiently exploit representational building blocks in learning more complex tasks: Instead of, for instance, learning behavioral policies via plasticity across all levels of a deep neural network that goes from the retina to action selection systems, hence dealing with the curse of dimensionality inherent in mapping sensory inputs and internal states onto motor outputs, the system can exploit stable sensory and motor representations.[29,57,145] Similar mechanisms are implemented in rudimentary form in continual learning and AI model training, e.g. when earlier layers are frozen in later fine-tuning,[27] [29,30] learning rates are steadily reduced in annealing protocols,[146-148] or when largely different learning rates are applied to different, hierarchically organized system components.[149] The full potential of the brain's modular architecture for complexity reduction, with finely tuned developmental time courses of the different components, has not yet been tapped. Schemata and careful construction of reusable components which can be compositionally rearranged is likely part of the brain's success in quickly adapting.[29,67,68]

**Conclusions and future directions**

So far, AI models mainly made use of two classes of mechanisms for adaptation to novel domains and shifting environments: First, continual learning algorithms ultimately relying on classical GD-based in-weights learning, but incorporating specialized mechanisms like experience replay or Bayesian updating. These mainly prevent catastrophic forgetting while tracking distribution shifts.[20,33] On the other hand, the power of massively pre-trained foundation models enables to perform OOD inference in-context. This capability, however, seems to rely more on recombination of functions learned in training rather than meta-learning of novel functions.[46,48] Thus, the large training corpus and long context window of foundation models play a huge role here. This capacity for generalization is therefore resource-intense, and the ability to quickly adjust to novel circumstances will still be limited by what has been experienced in training. In contrast, the dynamical and plasticity mechanisms discussed here have hardly been acknowledged in modern AI.

In real-world environments, time is of the essence. Physical, chemical, biological, and societal processes evolve across timescales from milliseconds to years. They have inherent non-random dynamics, from simple oscillatory processes to complex, multi-level, hyper-chaotic dynamics governed by a plethora of interacting nonlinear feedback loops. Brains need to deal with such dynamically evolving environments for predicting future outcomes and preparing for changing conditions, and in some sense science is the culmination of this process in humans. This capacity is reflected in the flexible dynamical mechanisms engaged by brains for representation, computation, prediction, and fast adaptation, as reviewed in the section on neuro-dynamical mechanisms above. It is supported by a plethora of biophysical and biochemical time constants and timescales, with cellular processing of inputs spanning milliseconds to minutes and hours. Furthermore, these timescales systematically differ among brain areas, from vision to cognition, to accommodate specific computational needs.[112,150-153] How closely brain dynamics tracks dynamics in the external world is perhaps most evident in various resonance phenomena, from the quick synchronization of hand clapping in a theater to the entrainment of brain oscillations by music or flickering lights, to the degree that epileptic seizures could be evoked.[154-156] In the real world, not only predicting an event, but also the timing of that event, is crucial.[96,157] Often organisms need to trade off the capability to respond fast, e.g. to a threat, versus launching an optimal response. Time is of the essence, and it is no coincidence that computational neuroscience has adopted dynamical systems theory as one of its major theoretical workhorses.

We therefore believe that AI research could profit tremendously from paying more attention to dynamical mechanisms that operate on multiple time-scales. They not only provide sensible priors for our dynamical world, but also computational scaffolds for quickly adjusting to changing conditions. This is particularly important for AI agents like robots or self-driving cars interacting in real time with the physical world. It is also essential for AI models for time series prediction that need to forecast non-stationary processes possibly crossing tipping points, like trajectories of patients suffering from bacterial infections or degenerative conditions, or climate systems.[158,159]

Until now, most AI models start from a global loss function that is iteratively optimized through GD or other gradual numerical procedures. This process is costly, prone to catastrophic forgetting, and often simply too slow to keep up with rapidly changing environments and unexpected events. While several useful strategies have been advanced for alleviating catastrophic forgetting, the other two issues are still largely unresolved. Rapid, one-shot forms of synaptic plasticity, or, more generally, multiple plasticity time scales adapted to the time scales of change in natural environments, may equip agentic AI with the solutions. BTSP is not only rapid, but may also be particularly resource-efficient,[2,3] endowing brains with long context-windows not suffering from the quadratic scaling with sequence length that bugs transformers.[160]

Cortical memory systems are efficient and resource-saving in various ways: This starts with the fact that information is filtered for importance and relevance by attentional mechanisms, partly controlled by neuromodulatory and motivational circuits.[161] Complementary memory mechanisms allow for storing everything away quickly while novel experiences are made, and then offline selectively embedding relevant information into the existing memory structure without erasing other crucial contents.[60,132,133] Unsupervised forms of learning are particularly relevant here; as we walk through our worlds in a continuous stream of sensory information, most often there is no explicit teacher, potentially not even explicit reinforcement signals, and what is really relevant and why, may only become apparent much later. Memory undergoes constant restructuring[2,120] that may serve abstracting generalizable schemata and principles or forming categories to compress information.[66,68,162,163] Hence, memory and information storage in the brain is selective, volatile, flexible, organized for purpose, and actively controlled by multiple loops, very different from how information is imprinted into current AI systems.

A much tighter integration of ML/AI and neuroscience research could speed up knowledge transfer and would benefit, we believe, neuroscience as much as it would help developing novel AI algorithms. In neuroscience, AI could lend a systematic, functional-computational perspective that could guide research questions in a specific way. For instance, research about principles of synaptic plasticity may incorporate ideas about potential optimization objectives that would yield specific testable hypotheses. A much tighter alignment of behavioral and cognitive task designs in both areas could allow for a more direct translation of findings at the neurophysiological level into AI architectures. For example, ML/AI algorithms for continual learning have so far been benchmarked mostly with incremental classification tasks, like task-incremental CIFAR (e.g.[33,34]). It is not clear whether these are actually the most useful testbeds for the real-world scenarios for which continual learning procedures are designed. Task designs used in animal research may offer more realistic and ecologically appropriate use cases, especially relevant for agentic AI. Vice versa, benchmarks in ML/AI are created to probe specific functionalities needed to advance methods. Hence, how animals would solve similar tasks could give insight into neural mechanisms that underlie precisely these computational functions. For example, in testing foundation models for time series forecasting,[164,165] initial segments of a time series are provided as context, and the model is then tasked with predicting the future evolution of the time series. In Rajalingham et al.[166], monkeys and humans see the trajectory of an object that then disappears behind an opaque screen, and the animal needs to predict at which spatial position it will reappear. This could easily be adjusted to probe time series forecasting abilities in animals. Tracing out a predicted trajectory with a cursor could be another idea for translating such task paradigms.

A helpful technology for directly translating between neuroscience and AI could be recent AI tools for *dynamical systems reconstruction (DSR)*: DSR models learn from a set of observed time series, like multiple single-unit recordings and concurrent behavioral responses, a *generative surrogate model* of the underlying dynamical processes (Fig. 5).[118,149,167] In principle, DSR models could be any type of NN architecture, like RNNs or auto-regressive transformers, embedded within modality-specific encoder-decoder architectures.[118] These models are then trained by specific control-theoretic procedures that help to capture long-term statistical and geometrical aspects of the dynamics.[81,148,168] Using such techniques, any modern AI architecture may be trained on neurophysiological activity while animals perform specific cognitive tasks, thereby directly inheriting the underlying neuro-dynamical and computational principles for solving such tasks from the data. Likewise, ML methods for inferring synaptic plasticity rules directly from neurophysiological and behavioral observations have been advanced recently,[169-171] without the detour of explicit computational model construction and subsequent experimental testing. These methodologies may help to tighten the link between ML/AI and neuroscience research, and for integrating neuroscience data more directly into AI model and training algorithm design.

## Acknowledgements

D.D. appreciates funding from the German Science Foundation (DFG) via grants Du 354/14-1 (within the research cluster FOR-5159 dedicated to prefrontal flexibility) and Du 354/15-1. BA was supported by the Intramural Research Program of the NIMH (ZIA MH002928-01). G.K. acknowledges funding from the Hector II foundation.

## Figures

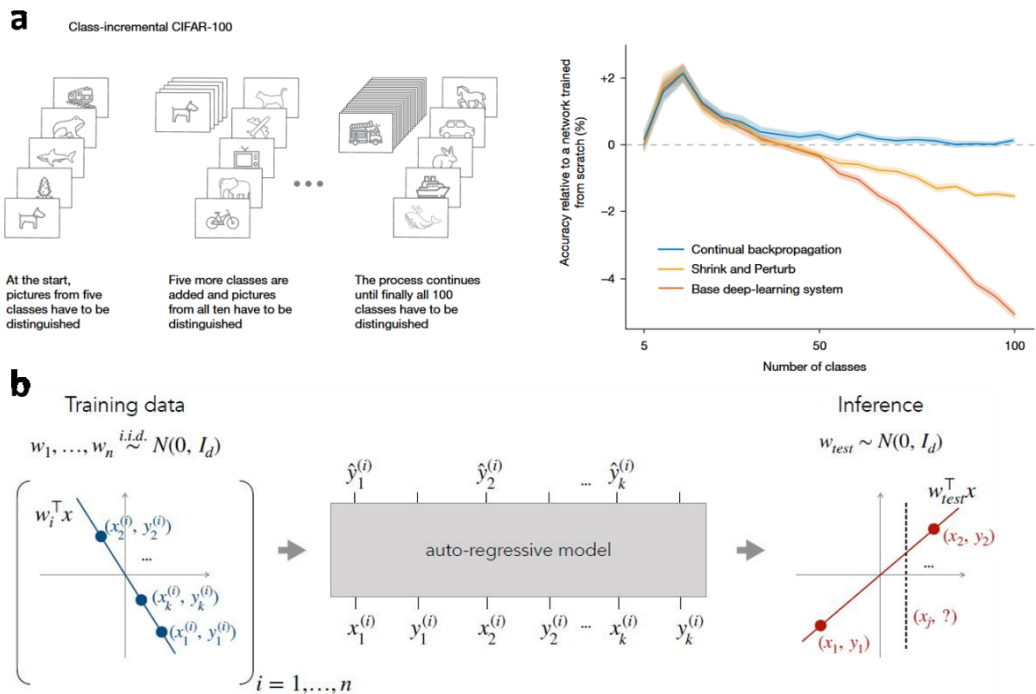

**Fig. 1. Continual and in-context learning.** a) Left: In class-incremental CIFAR, new image classification tasks are continuously added to the previous repertoire of already learned classifications. Right: For common deep learning systems, performance steadily decays below that of networks trained from scratch as new classification tasks are kept on being introduced. Continual backpropagation is one recently introduced training algorithm to amend this problem. <u>Reproduced from Dohare et al. (2024) with permission from the authors.</u>[33] b) In-context learning refers to the observation that large foundation models can generalize to new function classes without any model retraining, just by being presented with examples from a new function "in-context". In this example, an auto-regressive transformer model (center) was trained on pairs from many regression functions (left), then presented with a series of examples from a new regression function not contained in the training set (center), and asked to predict outputs from that new function given specific inputs (right). Surprisingly, the model can do this almost as well as a statistical regression model trained from scratch on the new pairs of input/ output values, but without any new parameter adjustment. <u>Reproduced from Garg et al. (2022) with permission from the authors.</u>[9]

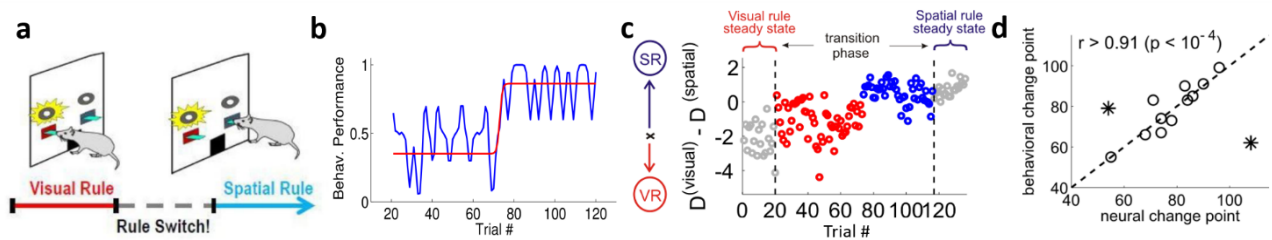

**Fig. 2. Sudden transitions in behavior and neural population representations during rule learning.** a) Animals are first trained on a 'visual' operant conditioning rule, and then – unknowingly – switched to a different rule in which the previous cue needs to be ignored and instead place (lever site) becomes relevant for reward (i.e., task contingencies change and the animal has to infer this by trial and error). b) At some point after the experimental rule switch, performance of individual animals rather abruptly jumps from chance level to nearly perfect performance (gradual learning curves as often seen in the literature are an 'artifact' of averaging across many such jumps[69]). c) These transitions in behavior are accompanied by rather abrupt transitions in the neural population activity. The trials at which these rapid transitions are observed can be statistically inferred through change point analysis, and change points observed in neural activity and behavioral performance tightly correlate. <u>Reproduced from Durstewitz et al. (2010) with permission.</u>[71]

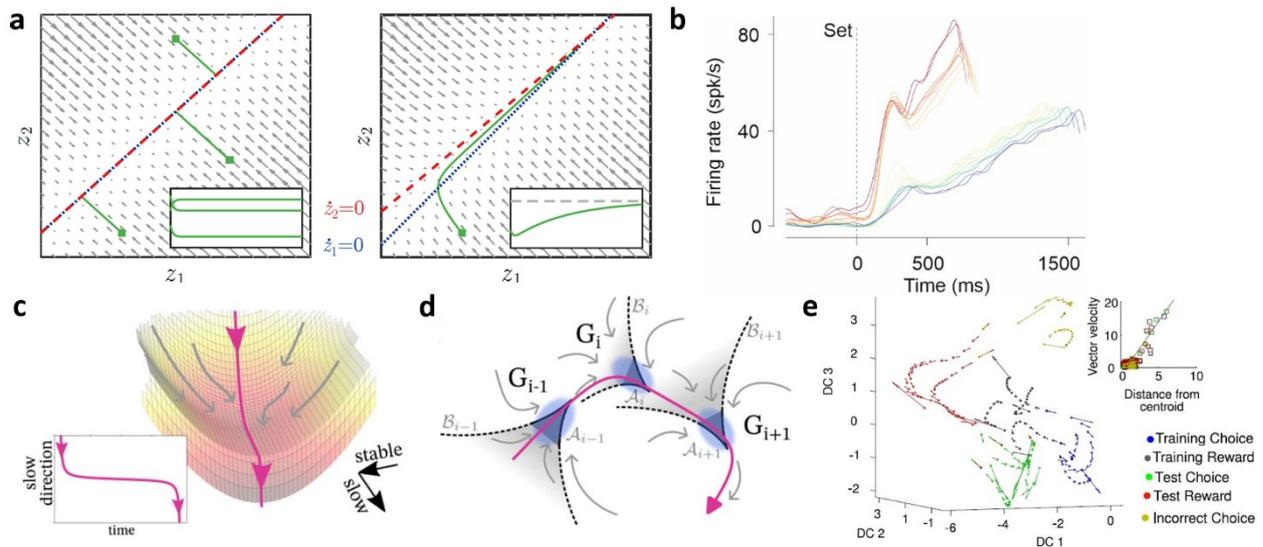

**Fig. 3. Dynamical mechanisms and their experimental support.** a) Manifold attractors (left), here illustrated in a 2d state space spanned by dynamical variables $z_1$ and $z_2$, are continuous sets of marginally stable fixed points, toward which the flow is directed from all sides as indicated by the system's vector field in gray (which gives the direction and magnitude of change at each point in state space as determined by the system's differential equations). Formally, fixed (or equilibrium) points are points in state space at which the flow of all state variables vanishes, i.e. the velocity vector becomes zero, indicated by the overlapping blue and red lines (the so-called nullclines). Trajectories (green) from various initial conditions (squares) converge to a point on the manifold attractor, which thereby stores a memory of the initial condition. Insets are time graphs. As the manifold attractor is slightly detuned (right graph), i.e. nullclines for the different state variables are not precisely aligned anymore, flow velocity continuously changes from zero, creating regions of arbitrarily slow flow. This phenomenon can be used to create arbitrary timescales in the system along a "ghost manifold". Reproduced from Schmidt et al. (2021).[102] b) Ramping firing rates flexibly adjusting to the length of a temporal interval between different events (in this case a set signal and a motor response), as observed in various brain areas,[111,172] are consistent with the idea of such ghost manifolds. Reproduced from Wang et al. (2018)[172] with permission from the authors. c) A ghost attractor is an attractor which just lost stability, such that in its vicinity the flow is arbitrarily slow (as indicated by the time graph in the inset, and similar to the right hand side in a). Trajectories enter the ghost attractor region (with the flow visualized by a pseudo-potential) from almost all directions but slowly leave along one direction again. d) Different ghost attractors $G_i$ may be connected in a ghost attractor chain, a construction that can flexibly implement sequences. c-d: Reproduced from Koch et al. (2024) with permissions.[109] e) Experimental support for such constructs in the nervous system comes from neural recordings during multiple item working memory and decision making tasks. Shown is a state space reconstructed from multiple single unit recordings with estimated flow vectors along trajectories, illustrating slowing down (quantified in the inset) as the system's state enters different task stages (color-coded). Reproduced from Lapish et al. (2015).[110]

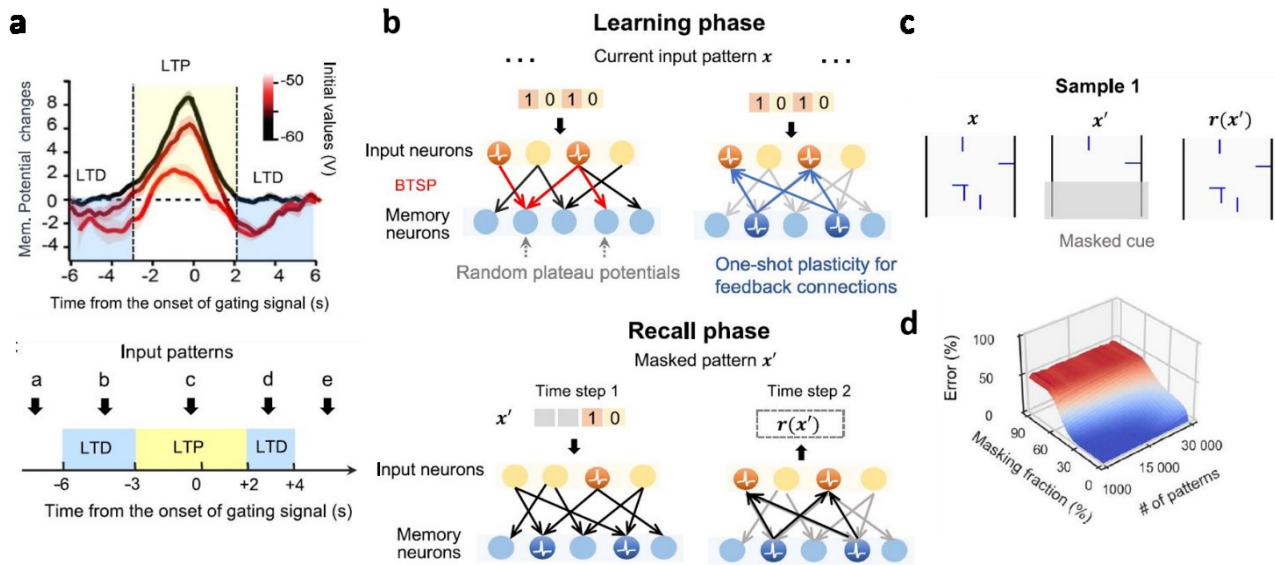

**Fig. 4. Fast plasticity mechanisms supporting episodic memory.** a) Behavioral time scale plasticity (BTSP; top) and its simplified formalization (bottom). BTSP depends on both, the temporal lag between synaptic inputs and a dendritic plateau potential (x-axis), and the initial strength of the synaptic weight (color coding), with a weight increase (long-term potentiation, LTP) induced if synaptic input and dendritic depolarization coincide within a time window of a few seconds, and a weight decrease (long-term depression, LTD) otherwise. b) Top: By implementing such a BTSP mechanisms in a 2-layer network, in which dendritic plateau potentials are induced in memory neurons at random times, different patterns can be instantaneously stored. Retrieval (completion of an original input pattern at the input layer) is accomplished by one-shot Hebbian (correlational) learning of feedback connections (right hand side). Bottom: Through these feedback connections, the originally observed pattern can be fully restored on the input layer from partial (occluded or noisy) cues, leading to memory retrieval similar as in a Hopfield network. c) Example of recovery, $r(x')$, of the original input pattern $x$ from a partially occluded cue $x'$. d) Error rate for downstream classification of recalled patterns as a function of memory load (# stored patterns) and the fraction of input bits masked. <u>Reproduced from Wu & Maass (2025) with permission from the authors.</u>[3]

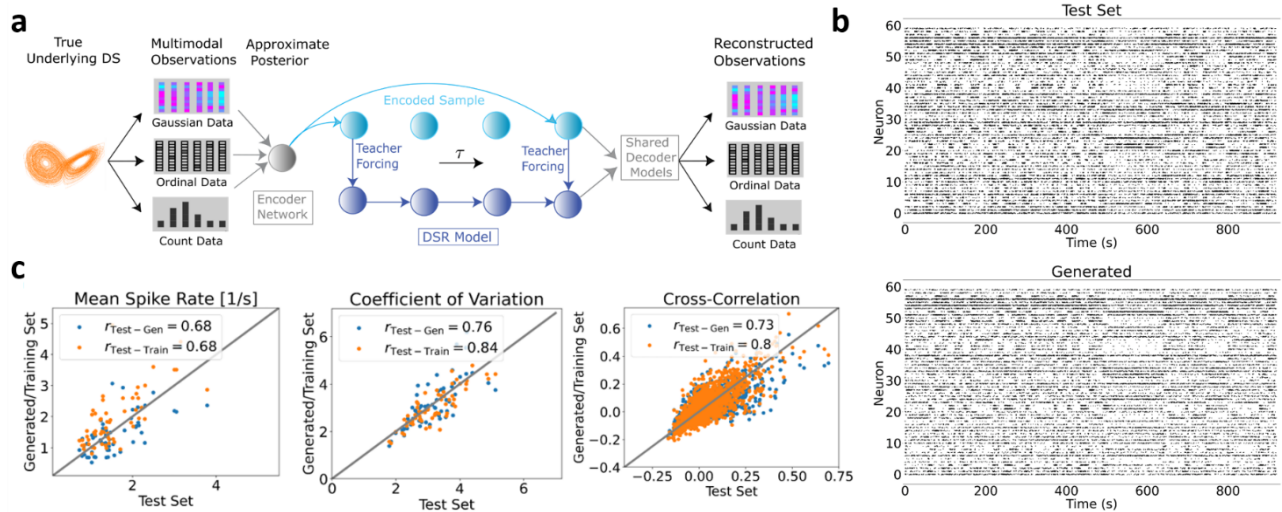

**Fig. 5. Dynamical systems reconstruction: Learning computational surrogate models from neuronal and behavioral data.** a) Multimodal teacher forcing (MTF) architecture. In MTF, a dynamical systems reconstruction (DSR) model, like an RNN, is controlled in training by a multimodal teacher signal derived from an encoder network (like a temporal CNN) that integrates across different measurement modalities with different statistical properties (like spike count and ordinal behavioral response data). The DSR model and the multimodal variational autoencoder are coupled to the same set of modality-specific decoder models (with same parameters), which map the latent states onto the observations, ensuring consistency between their latent codes. b) Examples of true (top) and DSR model-simulated (bottom) spike trains on an independent test set, from which only the initial condition was inferred to seed the DSR model. c) Agreement in different spiking statistics between true and model-generated spiking activity (blue) is about as good as between different segments of the real data (orange), indicating the model became a faithful surrogate of the real system within the bounds of statistical fluctuations. <u>Reproduced from Brenner et al. (2024).</u>[167]